\documentclass{article}
\usepackage{iclr2027_conference,times}

\usepackage{amsmath,amsfonts,bm}

\def\eqref#1{equation~\ref{#1}}
\def\1{\bm{1}}

\DeclareMathAlphabet{\mathsfit}{\encodingdefault}{\sfdefault}{m}{sl}
\SetMathAlphabet{\mathsfit}{bold}{\encodingdefault}{\sfdefault}{bx}{n}

\iclrfinalcopy

\usepackage{hyperref}
\usepackage{url}
\usepackage{graphicx}
\usepackage{booktabs}
\usepackage{multirow}
\usepackage{amssymb}
\usepackage{xcolor}
\usepackage{tikz}
\usetikzlibrary{positioning,calc,arrows.meta,decorations.pathreplacing,patterns}
\usepackage[capitalize,noabbrev]{cleveref}
\usepackage{enumitem}
\usepackage{microtype}
\usepackage{float}

\definecolor{cblue}{HTML}{2A78D6}
\definecolor{corange}{HTML}{EB6834}
\definecolor{cgray}{HTML}{8A8984}
\definecolor{cred}{HTML}{E34948}
\hypersetup{colorlinks=true, linkcolor=cblue!80!black, citecolor=cblue!80!black, urlcolor=cblue!80!black}

\newcommand{\vopen}{\texttt{OPEN}}
\newcommand{\voponly}{\texttt{OP\_ONLY}}
\newcommand{\vblocked}{\texttt{BLOCKED}}
\newcommand{\code}[1]{\texttt{#1}}
\newcommand{\sx}{\textsf{sx}}
\newcommand{\sy}{\textsf{sy}}
\newcommand{\sz}{\textsf{sz}}
\newcommand{\sw}{\textsf{sw}}
\newcommand{\ci}[2]{\,{\footnotesize[#1,\,#2]}}

\title{Written as a Record, Read as an Address: What a Forward Pass Leaves in an Operation's KV Cache}

\author{Lingfeng Wu\textsuperscript{1}, Behzad Shomali\textsuperscript{1, 2}
\\[6pt]
    \textsuperscript{1}University of Bonn \
    \textsuperscript{2}Lamarr Institute \ \\[6pt]
    \texttt{s64lwu@uni-bonn.de}
}

\renewcommand{\headrulewidth}{0pt}
\begin{document}

\maketitle

\begin{abstract}
    When a language model reads an operation such as \emph{``Swap the contents of Box F and Box B''}, its forward pass writes keys and values for those tokens into the KV cache. Prior work on entity tracking establishes what models \textit{use}: bindings are resolved at query time rather than stored as explicit latent state. We ask what they \textit{write} at the operation span and how it is accessed. We split a forward pass into a frozen writer and a reader: the writer's cache is recomputed without gradients, while the reader sees only the instruction and operation tokens, with all state descriptions hidden, and is trained in isolation. Anything the reader recovers was therefore already present in the unmodified cache. On a synthetic boxes task, a base reader recovers $\leq 0.06$ of queried bindings against $0.75$--$1.00$ after training, and recoverability tracks the operation's read/write footprint. We find two modes of access. Across Llama-3.1-8B and Mistral-7B, operation-span transplants causally redirect which visible state is read even when the two worlds hold identical values, revealing a \textit{routing} record. Isolation training preserves routing and adds direct access to the \textit{payload}, the value the operation read, from the single operand-name token in a narrow mid-depth band (layers 12--15 of 32 in Llama-3.1-8B, 14--17 in Mistral-7B) --- the same site that holds the routing record. The same recipe extends to further operations, ToMi and GSM8K, but is bounded by training coverage and costs open-book accuracy. Operation tokens thus leave localized, causally recoverable records that support both routing and direct payload access, though the model that writes them reads mainly the address they carry and not the value.

\end{abstract}

% =====================================================================
\section{Introduction}
\label{sec:intro}

Language models track how a described world changes as they read: which box now holds the comb, or what a variable holds after an assignment. In a decoder-only transformer, information moves between positions only through the key/value (KV) cache, so whatever later computation knows about an update must either be stored in the cache entries written while reading it or be recomputed when a question arrives. Prior work on entity binding and tracking mostly finds the latter: bindings are resolved at query time rather than updated eagerly \citep{kim2023entity,feng2024binding,prakash2025lookbacks,oh2026rebinding}, and models rebuild state from visible tokens instead of maintaining it incrementally \citep{tang2026entity}.

These findings concern what the model \emph{uses}; they leave open what it \emph{writes}. Probing alone cannot settle the question, because a probe can decode information that the model never uses \citep{hewitt2019control,elazar2021amnesic,belinkov2022probing}, and patching can show that a site matters without showing what it contains \citep{vig2020causal,geiger2021causal,zhang2024patching}. To measure the gap between what is written into the cache and what the model reads from it, we hold the writer fixed and vary only how the cache is read.

\begin{figure}[t]
\centering
\includegraphics[width=\linewidth]{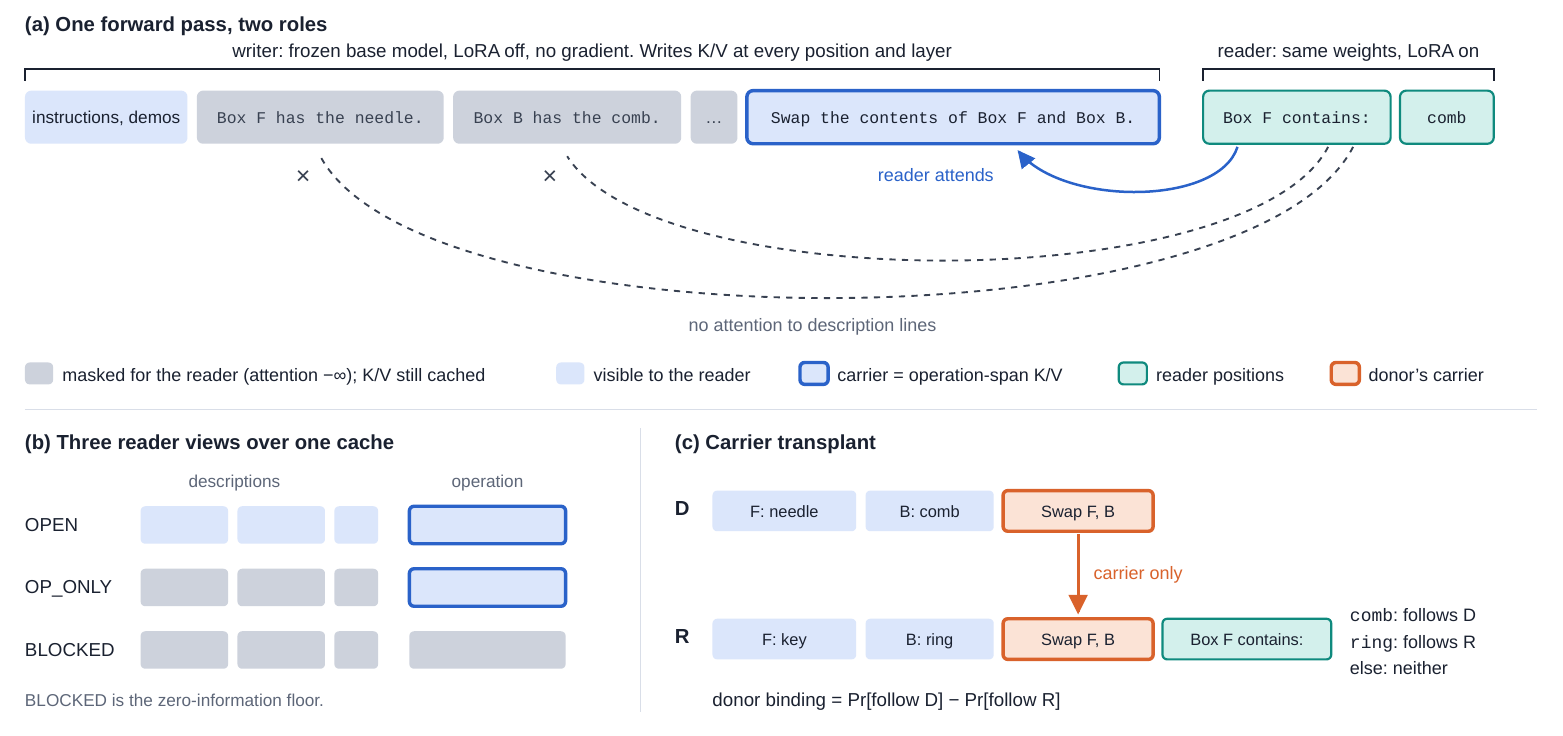}
\vspace{-15pt}
\caption{\textbf{Setup, shown on one boxes example}. Anything the reader answers under \voponly{} must come from the operation span's K/V. Color encodes what the reader may attend to: gray spans are masked for the reader (attention logits $-\infty$; their K/V remain in the cache), blue spans are visible, and the outlined span is the carrier, i.e. the operation-span K/V. (a) The prefix is processed once by the frozen base model (writer, LoRA adapter off). The reader is the same network at the query and answer positions with the adapter on. The operation line names two box letters and no item. (b) Three views over one frozen cache. (c) A transplant copies only the carrier of a donor world D into a recipient world R that differs only in its items; the answer then follows D, follows R, or neither.}
\label{fig:schematic}
\end{figure}

\paragraph{Frozen writer, trained reader.} Each prompt consists of a set of state descriptions, one operation statement and a query, each occupying a contiguous span of token positions. For simplicity, we refer to these token spans as lines: the spans that state the current bindings are description lines (in code, assignments such as \code{a = 3}) and the span that changes them is the operation line. We mask the reader's attention to every description line and leave the operation line visible (\cref{fig:schematic}). This line names no item, but its K/V were computed after the writer had processed the description lines, so they can depend on the items those lines mention. We then train a low-rank adapter \citep{hu2022lora} on the reader alone. The writer's weights never change and its cache is recomputed without gradients at every step, so anything the reader recovers is recoverable from the unmodified cache. This does not imply an explicit state variable or tell us whether the adapter performs a lookup or a new computation (\cref{sec:limitations}). We study two tasks with this structure, a natural-language boxes task (\cref{fig:schematic}) and a code task, which lets us control exactly which variables an operation reads and which it writes. We find that the two questions come apart: the base model already uses the operation span to decide where to read, but barely recovers what it holds. Isolation training exposes the second use at the same site as the first. Our contributions are as follows:

\begin{itemize}[leftmargin=1.2em, itemsep=1pt, topsep=2pt]
\item \textbf{Storage without native retrieval} (\cref{sec:storage}). We show that the operation span stores information that the base model rarely retrieves. With every description line hidden, the base reader answers at most $0.06$ of the boxes queries on three models (Llama-3.2-1B, Llama-3.1-8B, Mistral-7B), whereas a reader trained in isolation answers $0.75$--$1.00$ of them.
\item \textbf{Operation-local footprint} (\cref{sec:footprint}). We show that what can be recovered follows the operation's read/write footprint. For an assignment such as \code{a = b + 1}, the values of the	written variable \code{a} and of the read-only operand \code{b} become recoverable from the operation span, while the values of variables the operation does not mention do not.
\item \textbf{Two uses of one cache} (\cref{sec:interfaces}). The base model uses the carrier to select which visible line to read, even when donor and recipient hold identical values. Isolation training does not weaken this routing and adds access to operation-local values. Both results hold for three reader seeds, and both routing and direct payload replicate in Mistral-7B.
\item \textbf{Where the payload is read} (\cref{sec:localize}). The trained reader reads it from the K/V of a single token, the operand name (\code{b} in \code{a = b + 1}), in layers 12--15 of 32, with a weaker contribution from 8--11 (14--17 in Mistral-7B). This is the site where the base model already keeps its routing record, so training exposes an existing record rather than creating a new store.
\item \textbf{Breadth and limits} (\cref{sec:breadth}). With each reader trained from scratch, the same recipe transfers to four non-literal code operations, ToMi and GSM8K, while ordinary fine-tuning on the same data stays at the base level; access is bounded by training coverage and costs open-book accuracy.
\end{itemize}

% =====================================================================
\begin{figure}[t]
\centering
\includegraphics[width=\textwidth]{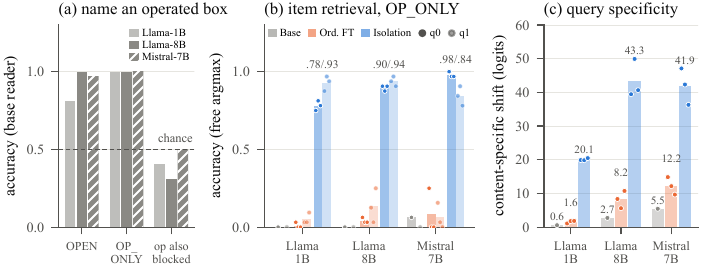}
\vspace{-15pt}
\caption{\textbf{Boxes task on three models} (Llama-3.2-1B, Llama-3.1-8B, Mistral-7B; 32 families each). \textbf{(a)} Mask validity: under \voponly{} every base reader names an operated box perfectly, and it falls to about chance when the operation is also blocked. \textbf{(b)} Free-vocabulary retrieval of the item now in the queried box under \voponly{}. Bars are means over three training seeds and dots are individual seeds; the base readers have no training seed. \textbf{(c)} Query specificity: the logit shift toward a donor's correct answer after a carrier transplant, minus the shift caused by a query-irrelevant donor. Dots are seeds.}
\label{fig:boxes}
\end{figure}
\section{Method}
\label{sec:method}

\paragraph{Writer, carrier, reader (\cref{fig:schematic}a).} An instance consists of a fixed prefix, description lines $s$ (the box's content, or the assignment in code), one operation line $o$ and a query $q$. A \emph{world} is a complete setting of $s$. The \textbf{writer} is the base model with the LoRA adapter disabled. It processes the prefix, $s$ and $o$, and leaves K/V at every position and layer. The \textbf{carrier} is the writer's K/V at the positions of $o$, across all layers. We call the information causally recoverable from the carrier its \textbf{record}. The \textbf{reader} is the forward pass at $q$ and the answer tokens. It shares all weights with the writer and differs only by an activated LoRA adapter.
The adapter is applied only at the query and answer rows, so it never alters the cache it reads. We say that the cache stores information when that information is causally recoverable in this sense. No information can therefore enter the cache through the training signal; training can change only how an existing cache is read.

\paragraph{Views (\cref{fig:schematic}b).} We experiment with three different views. A view sets the reader's attention logits to $-\infty$ at a set of prefix positions. \vopen{} blocks nothing. \voponly{} blocks every line of the instance except $o$, including all description lines. \vblocked{} blocks the whole contiguous instance, separators included, and defines the zero-information floor.

\paragraph{Isolation training.} The reader carries a rank-16 LoRA on \code{q/k/v/o} and is trained with cross-entropy on the answer tokens. Its prefix cache is recomputed at every step by the base model with the adapter disabled (full recipes in \cref{app:training}). Two arms share data, seeds, steps, optimizer and pair schedule and differ \emph{only} in which prefix positions the reader may see during training: \textbf{isolation training} (ISO) trains under \voponly{}, and \textbf{ordinary fine-tuning} (ORD) blocks nothing. ORD controls for the additional optimization and answer supervision.

\paragraph{Transplants (\cref{fig:schematic}c).} A \emph{pair} is two worlds that differ only in assigned values. Within a pair, a transplant replaces the carrier of a \textbf{recipient} world $R$ with that of a \textbf{donor} world $D$ at every layer, after asserting identical span positions and prefix lengths. A \emph{matched} donor changes the value that the queried variable ends up holding. \emph{Mismatched} and \emph{irrelevant} donors are controls. 

\paragraph{Metrics.} Besides exact-match accuracy we report \textbf{donor binding}, $\Pr[\text{answer follows }D]-\Pr[\text{answer follows }R]$; \textbf{redirection}, the rate at which the answer follows a different recipient-visible source; and \textbf{payload readout}, the fraction of answers equal to the donor's value when every description line is masked.

% =====================================================================
\section{Experimental setup}
\label{sec:setup}

\paragraph{Tasks.}
\emph{Boxes}: description lines are such as \code{Box F has the needle.}, the
operation is \code{Swap the contents of Box F and Box B.} and the query is \code{Box F contains:}. Each instance carries two queries, one per operated box: $q_0$ asks about the box named first in the operation and $q_1$ about the second. We evaluate 32 item families (a family fixes the items and box letters from which a pair of worlds, both questions and all donors are built), and score the full-vocabulary argmax at the first answer token.

\emph{Code}: four assignments with random names and values from 10--25, an operation line that contains names only (\code{a, b = b, a} or \code{a = b} or \code{a = b + 1}) and the query \code{\# print(a) ->}, scored by exact match of greedy generations. The swap is queried like the boxes task, $q_0$ for \code{a} and $q_1$ for \code{b}. Unlike a swap, \code{a = b} writes \code{a} and only reads \code{b}, so we instead query four roles: the write target $\sx$, the read-only operand $\sy$, and two unmentioned variables, $\sz$, which holds the same value in both worlds of a pair, and $\sw$, which does not. All values are single tokens, so the two worlds of a pair are position-aligned, as a transplant requires.

\paragraph{Models and training.}
\emph{Boxes} uses Llama-3.2-1B-Instruct \citep{grattafiori2024llama3} in FP32 and, with the same pairs, schedules and seeds, Llama-3.1-8B-Instruct and Mistral-7B-Instruct-v0.3 \citep{jiang2023mistral7b} in NF4 \citep{dettmers2023qlora}; Mistral prompts are re-rendered with its own chat template (\cref{app:training}). Code uses Llama-3.1-8B-Instruct in NF4, because the 1B model has a low base accuracy (\cref{app:extra,app:localize}). Mistral-7B replicates routing, payload, the payload site and the held-out footprint (\cref{sec:footprint,sec:interfaces,sec:localize}).

% =====================================================================
\section{Results}
\subsection{Storage without native retrieval}
\label{sec:storage}

All results here use \voponly: boxes on three models (\cref{fig:boxes},
\cref{tab:boxes}) and unseen code instances on the 8B model.

\paragraph{Base reader.}
\Cref{fig:boxes}a checks the mask: it hides the description lines but not the operation, so the base model still names the operated boxes in 32/32 families and drops to about chance (0.41) once $o$ is also blocked. The answer item never appears in the visible text, and the base reader's free-vocabulary accuracy is 0.000 on both questions ($q_0$, $q_1$; \cref{fig:boxes}b). The span is not ignored: a donor's carrier shifts the base model's logits toward the donor's answer in all 32 families ($S_{\text{BASE}} = 1.496$, 95\% CI $[1.16, 1.86]$; \cref{app:native}), but changes the generated answer in only 1 of 32. The base model reads the span, but not reliably enough to answer from it. 

\paragraph{Isolation training.}
\Cref{fig:boxes}b and \cref{tab:boxes} give the effect of training. Over three seeds, isolation training reaches 0.781 on $q_0$ and 0.927 on $q_1$, and its training loss falls from 1.90--2.00 to 0.30--0.37 (first vs.\ last 64 of 256 steps). Ordinary fine-tuning stays near zero on the 1B model ($q_0 \le 0.031$, $q_1 \le 0.094$) and at or below $0.25$ in every seed of the two larger models (\cref{tab:boxes}), although it saw the same pairs in the same order, so what matters is whether the description lines were visible during training. \Cref{fig:boxes}c shows that the readout is query-specific: relative to a query-irrelevant donor, a donor's carrier moves the answer logit toward the donor's answer by 0.57 for the base model, 1.09--1.84 for ORD and 19.86--20.49 for ISO on Llama-3.2-1B, with the same ordering at larger magnitudes on the 8B and Mistral readers (\cref{fig:boxes}c).

\paragraph{Unseen instances.}
We repeat the comparison on code swap with the 8B model, on 300 unseen items from an unused
seed. Under \voponly{} the base reader scores $0.152$\ci{.12}{.18}, ordinary fine-tuning $0.112$/$0.120$/$0.142$ and isolation training $0.627$/$0.688$/$0.610$ over three seeds, with ordinary fine-tuning at or below the untrained reader throughout. Moreover, the effect is specific to the operation span: a later filler line, whose K/V attended to strictly more of the prompt, supports only $0.055$, and a donor whose values all lie outside the item reduces the trained readers to $0.013$--$0.040$.

\subsection{The operation's local footprint}
\label{sec:footprint}
\begin{figure}[t]
\centering
\includegraphics[width=\textwidth]{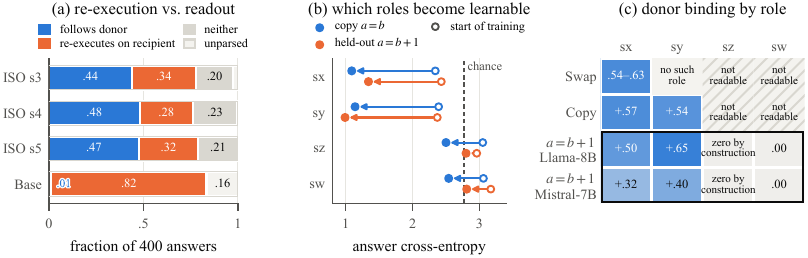}
\vspace{-15pt}
\caption{\textbf{Accessible information tracks the footprint (code; Llama-3.1-8B unless a row names another model).} 
\textbf{(a)} Swap under \vopen{}: the base reader re-executes on the recipient's visible state, trained readers return the donor's realized value (\cref{app:carrier}). \textbf{(b)} Answer cross-entropy from the start of isolation training (open) to its end (filled), chance $\ln 16$. \textbf{(c)} Donor binding under \voponly{}. Only the addressed roles move and bind. A swap has \emph{no such role} at \sy{}, \emph{zero by construction} marks \sz{} cells holding the same value in both worlds, and hatched cells are \emph{not readable} by that reader.}
\label{fig:footprint}
\end{figure}

A box query concerns a box that the swap both reads and writes, so this section and the next two use the code task (\cref{sec:setup}), which separates the roles.

\paragraph{Re-execution versus readout.} A carrier that encoded only a reusable operator could be \textbf{re-applied} to the recipient's state, and only \vopen{}, where that state remains visible, distinguishes this from a \textbf{record} of the realized result. \cref{fig:footprint}a shows that the base reader re-executes the operation on the recipient's visible state in $0.820$ of answers and returns the donor's realized value in only $0.013$: the base model uses the carrier mainly to identify \emph{which} operation to apply. The three isolation-trained seeds instead return the donor's realized value ($0.438$/$0.480$/$0.475$, against $0.340$/$0.282$/$0.315$ re-execution), so the trained readers trust the donor's computed result rather than re-deriving it.

\paragraph{Learnable roles.} In \code{a = b}, \code{a} is written and \code{b} is only read. \Cref{fig:footprint}b tracks what isolation training makes learnable: under isolation training, answer cross-entropy falls for \sx{} ($2.34\!\to\!1.09$) and \sy{} ($2.39\!\to\!1.14$) but only slightly for \sz{} and \sw{} ($3.05\!\to\!2.50$, $3.06\!\to\!2.54$; chance $2.77$), and under \vopen{} the trained reader answers $0.930$/$0.890$ on the addressed roles against $0.085$/$0.110$ on the others. 

\paragraph{Prediction on a held-out operation.} For \code{a = b + 1}, the footprint account implies that donor information is accessible for the \textbf{read set} $\cup$ the \textbf{write set}, here $\{b\}\cup\{a\}$. \Cref{fig:footprint}b and the boxed rows of \cref{fig:footprint}c test that prediction on an operation the account was not built on: the two addressed roles became learnable ($2.43\!\to\!1.34$, $2.37\!\to\!0.99$) while \sz{} and \sw{} stayed at chance ($2.80$, $2.81$), and donor binding under \voponly{} was $+0.500$ for \sx{} and $+0.650$ for \sy{}, against $0.000$ for the uninvolved frame variable \sw{} (\cref{fig:footprint}c; \sz{} holds the same value in both worlds of a pair, so its binding is zero by construction). A second  reader repeats this ($+0.615$, $+0.675$, $0.000$; \cref{app:carrier}). On this held-out operation, accessible information follows the read/write footprint. We treat this as a functional selectivity result; it does not show that the record is complete or discrete.

\subsection{Routing and payload}
\label{sec:interfaces}

Following \citet{prakash2025lookbacks}, we distinguish two functional interfaces. Through \textbf{routing} (addressing), the carrier determines which external source downstream computation reads. Through \textbf{payload access}, downstream computation recovers operation-local content without access to external state. We call a visible line a \textbf{causal source} of an answer if masking that line selectively removes the answer. Unless stated otherwise, probes in this section use Llama-3.1-8B with the operation \code{TARGET = OPER + 1} over four slots (TARGET, OPER, ALT, OTH), $n=120$. Here, TARGET and OPER are the \sx{} and \sy{} roles of \cref{sec:footprint}; ALT is an unmentioned variable that the donor reads instead, and OTH is read by no operation in either world.

\begin{figure}[t]
\centering
\includegraphics[width=\textwidth]{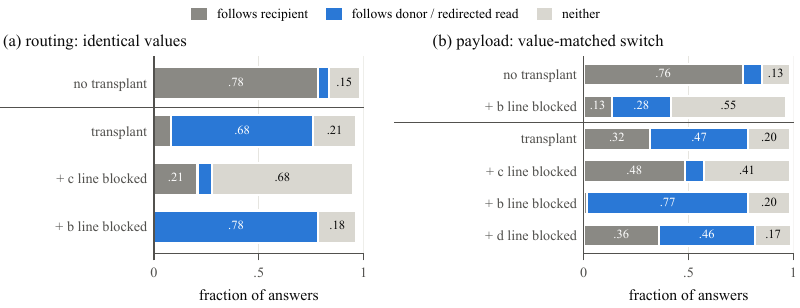}
\vspace{-15pt}
\caption{\textbf{Transplants redirect the read} (isolation-trained reader). \textbf{(a)} Routing: donor and recipient hold the same four values and differ only in which variable the operation reads (\code{a = c + 1} vs.\ \code{a = b + 1}). The read moves to the \code{c} line, and masking that line removes it; the untrained model behaves the same way. \textbf{(b)} Payload: the donor differs only in the operand's value (\code{b = 22}). After the transplant the answer moves from the \code{b} line to the visible line that carries \code{22}, and masking that line removes it.}
\label{fig:routing}
\end{figure}

\begin{figure}[t]
\centering
\includegraphics[width=0.62\linewidth]{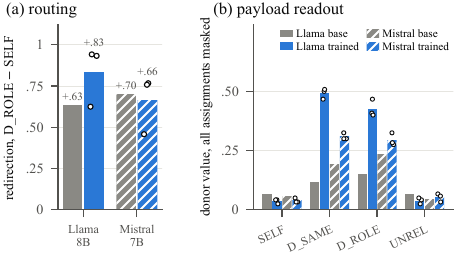}
\vspace{-10pt}
\caption{\textbf{Isolation training adds payload access and does not reduce native routing.} Both interfaces read the same carrier (Llama-3.1-8B and Mistral-7B, $n=120$ per reader; trained bars: mean of three seeds, dots: seeds). SELF is the recipient's own carrier, D\_SAME differs only in the operand's value, D\_ROLE keeps the values but reads another variable, and UNREL replaces all values. \textbf{(a)} Routing: D\_ROLE redirection minus SELF. \textbf{(b)} Donor value with all description lines masked.}
\label{fig:consume}
\end{figure}

\paragraph{Routing.} Take a recipient with \code{a = 14}, \code{b = 18}, \code{c = 20}, \code{d = 11} and the operation \code{a = b + 1} (answer \code{19}; OPER \code{b}, ALT \code{c}, OTH \code{d}). The donor holds the same four values but has the operation \code{a = c + 1} (a D\_ROLE donor), so only the named operand differs. \Cref{fig:routing}a follows where the answer comes from once the donor's operation-line K/V are transplanted into the recipient, the untrained model answers \code{21}, the value on the \code{c} line plus one, on $0.642$ of items against $0.008$ without a transplant ($+0.633$\ci{+.55}{+.72}), and masking \code{c = 20} reduces this to $0.000$. The isolation-trained reader behaves the same way ($+0.625$\ci{+.54}{+.71}; masking \code{c} gives $0.067$; \cref{fig:routing}a). The carrier therefore specifies which source to read rather than which number to output, and this routing is native. Two further reader seeds route even more strongly ($+0.942$, $+0.933$; \cref{fig:consume}a), and routing replicates in Mistral-7B (base $+0.700$; trained $+0.458$ to $+0.767$). Training therefore does not trade the native interface for the learned one.

\paragraph{The carrier addresses by value, not by name.} Now let the recipient have \code{a = 14}, \code{b = 18}, \code{c = 22}, \code{d = 11} and \code{a = b + 1} (answer \code{19}), and let the donor differ only in \code{b = 22} (answer \code{23}; a D\_SAME donor). Both operation lines name \code{b}, so a name-based address would still point to \code{b}, and in the recipient \code{22} appears only on the \code{c} line. \Cref{fig:routing}b tracks the answer as each line is masked in turn: after the transplant the three trained readers answer \code{23} on $0.467$/$0.300$/$0.508$ of items. Masking \code{c = 22} changes this by $-0.375$\ci{-.47}{-.29}/$-0.217$/$-0.325$ as the reader falls back to \code{b}, masking \code{b = 18} by $+0.300$/$+0.417$/$+0.342$, and masking \code{d = 11} by at most $0.050$. If no line holds \code{22}, donor-following is only $0.100$--$0.192$, and the untrained model shows none of this ($0.025$ in every cell). The reader thus uses the donor's value to find the visible line that carries it, so the carrier holds the value its operation read and not merely a name to re-resolve.

\paragraph{Direct payload.} With all four description lines \textbf{masked}, the reader sees only \code{a = b + 1} and can output \code{23} only from the carrier. \cref{fig:consume}b shows the trained readers do so on $0.467$/$0.500$/$0.508$ of items for the D\_SAME donor and on $0.408$/$0.467$/$0.400$ for a D\_ROLE donor that keeps the recipient's values and computes \code{a = c + 1} (again \code{23}), against at most $0.050$ with no transplant (SELF) or with an unrelated donor whose values are all replaced (UNREL; \cref{fig:consume}b). The carrier thus holds the value its operation read, from whichever variable, and the reader outputs it only when no description line is visible. Unlike routing, direct payload depends on training: the untrained model outputs the donor's value on only $0.117$ of items. In Mistral-7B the base model already shows a weak payload ($0.192$), which training raises to $0.300$--$0.325$ (\cref{fig:consume}b).

\paragraph{The native pathway is absent, not merely unused.} The base model is not ignoring the carrier---it routes with it, and by value---yet it almost never answers from it. Removing the visible description lines one at a time, with a donor whose value appears nowhere in the recipient, the trained readers climb from $0.100$--$0.192$ to $0.475$--$0.500$ while the base model stays at or below $0.025$ throughout; the decisive pair masks only the operand's line, leaving \emph{more} text on screen than masking the other three, yet only the trained readers gain payload there, and the base model's own accuracy falls from $0.933$ to $0.000$ without any fallback to the carrier (\cref{fig:ladder}). The record is there, but the base model uses it only as an address; what isolation training adds is a path from the same record to an answer (\cref{sec:localize}).

\subsection{The payload is read from the routing record}
\label{sec:localize}

\begin{figure}[t]
\centering
\includegraphics[width=\textwidth]{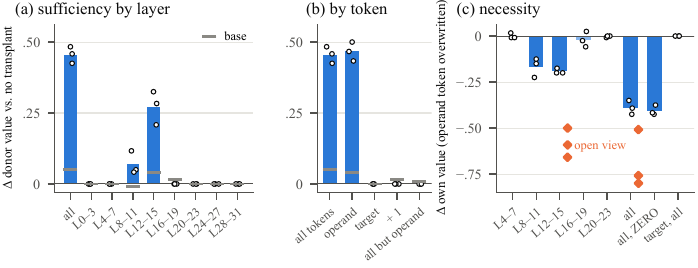}
\vspace{-20pt}
\caption{\textbf{The payload is read from the operand-name token in layers 12--15} (Llama-3.1-8B, all description lines masked, $n=120$ per reader; bars: mean of three trained readers, dots: readers, gray dashes: untrained model). \textbf{(a, b)} Sufficiency: donor-value readout when a D\_SAME carrier is transplanted only at one 4-layer band or only at some tokens, minus no transplant. \textbf{(c)} Necessity: change in own-value readout when the reader's operand-name K/V are overwritten (FRESH: state of a dangling name; ZERO: zeros); last bar: write-target token instead. Diamonds: description lines visible. Mistral-7B, where the site is layers 14--17: \cref{fig:payloc-mistral}.}
\label{fig:payloc}
\end{figure}

Isolation training adds payload access (\cref{sec:interfaces}), but where does the payload come from? The reader could use a new representation, for instance of the result at the tokens that complete the operation, or read a value that the base model already writes for routing. The routing record is a natural candidate: in the base Llama-3.1-8B it sits at the operand-name token in layers 12--15, where transplanting it redirects the read and overwriting it cuts own execution by $0.750$, and it encodes the operand's value along with its name and position (\cref{app:localize}). In all three trained readers, the payload is read from this record (\cref{fig:payloc}).

\paragraph{One token, holding the operand's value.} Transplanting only the operand-name token's K/V reproduces the whole-carrier readout ($0.475$/$0.508$/$0.525$ vs.\ $0.467$/$0.500$/$0.508$; no transplant $\le 0.042$), while every other token adds nothing. The same transplant makes the operand query return the donor's operand value ($0.467$--$0.558$), so the token carries the operand's value, and the addition is applied when the target is queried (\cref{app:payloc}). Conversely, we overwrite the reader's own operand token (\code{b} in \code{a = b + 1}) with its state from a prompt whose operation names an unassigned variable instead (\code{a = e + 1}; FRESH). This lowers the reader's readout of its own value by $0.35$--$0.43$, close to the floor, whereas the same overwrite of the write-target token (\code{a}) has no effect.

\paragraph{Which layers the payload is reading from.} Only layers 12--15 and, more weakly, 8--11 are sufficient ($+0.21$ to $+0.33$; $+0.04$ to $+0.12$) or necessary ($-0.13$ to $-0.23$ each; at most $-0.06$ elsewhere). With description lines visible, the same overwrite at layers 12--15 lowers own accuracy by $0.50$--$0.66$, so the state still serves as the address. The untrained model reads almost no payload from any part of the carrier (at most $+0.06$).

% =====================================================================
\section{Breadth and limits of learned access}
\label{sec:breadth}

\begin{figure}[t]
\centering
\includegraphics[width=\textwidth]{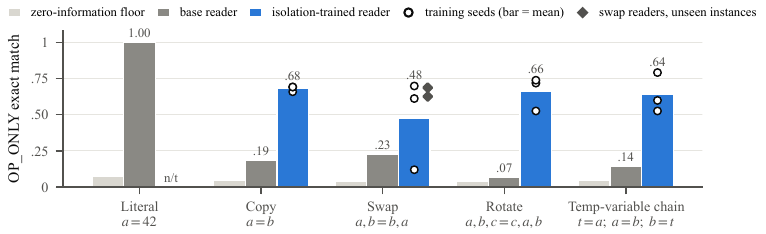}
\vspace{-15pt}
\caption{\textbf{Four non-literal operations and a visible-answer control} (code, 8B, \voponly{}, $n=150$). Each operation has its own freshly initialized reader. Bars are means of three training seeds (circles). Diamonds: the two swap readers of \cref{sec:storage} on unseen instances. The literal \code{a = 42} puts the answer on the visible line and is not trained (n/t).}
\label{fig:ops}
\end{figure}

\paragraph{Operations.} For the four non-literal operations the base reader is weak under \voponly{} ($0.067$--$0.227$). Isolation training raises accuracy to $0.527$--$0.793$ on 11 of 12 seeds, against a floor of $0.040$--$0.047$ (\cref{fig:ops}); the remaining swap seed reached $0.120$.

\paragraph{ToMi.} ToMi \citep{le2019tomi} stories contain one move line (\code{Jack moved the apple to the green\_box.}) and ask where the object was at the beginning, which only an earlier description line states. Treating the move as the operation line (Llama-3.1-8B, 768 steps, 200 validation questions), the base reader scores $0.040$ under \voponly{}, ordinary fine-tuning at most $0.040$, and isolation training $1.000$/$0.970$/$0.975$. Transplanted move lines bind the trained answer to the donor's start location ($\ge +0.985$) and are selective against donors that differ only in room names ($\ge +0.979$; \cref{app:nl}).

\paragraph{GSM8K and MMLU.} On GSM8K \citep{cobbe2021gsm8k} (Llama-3.2-1B), the writer reads the question and a reference solution, and the reader, blocked from the question and the solution's last line, must produce the final number. On the 176 items whose answer is not visible, isolation training scores $0.233$/$0.222$/$0.205$, against $0.011$--$0.023$ for ordinary fine-tuning and $0.023$ for the base reader (\cref{app:gsm8k}). The gap is not universal: on MMLU \citep{hendrycks2021mmlu} (Llama-3.1-8B) with the question masked, the base reader still scores $0.599$ (open $0.638$; \cref{app:nl}).

\paragraph{Limits.} Learned access collapses when values move from 10--25 to 30--45 ($0.030$/$0.040$/$0.168$ for three seeds; \cref{fig:boundary}a), a failure of the answer values rather than the context values: with out-of-range context but an in-range answer accuracy holds ($0.711$/$0.864$), and with in-range context but an out-of-range answer it collapses ($0.041$/$0.049$; crossed design, \cref{app:values}). Training on values 10--99 restores accuracy on 30--45 ($0.470$/$0.360$/$0.417$) but not on 200--299 ($0.015$/$0.025$/$0.000$; \cref{fig:boundary}b), so the boundary depends at least partly on training coverage. Access does not extend to unseen operations: a reader trained on three box operations answers ``nothing'' on a held-out fourth in 256/256 cases.

\paragraph{Cost.} Isolation-trained readers lose $0.105$/$0.168$/$0.017$ of \vopen{} accuracy on unseen code instances and more on GSM8K ($0.460$--$0.885$ vs.\ $0.970$; \cref{app:gsm8k}).

% =====================================================================
\section{Related work}
\label{sec:related}

\paragraph{Binding and entity tracking.} Models bind entities to attributes through binding IDs \citep{feng2024binding} and track entity states in context \citep{li2021implicit,kim2023entity}; fine-tuning strengthens existing tracking circuits rather than building new ones \citep{prakash2024finetuning}, and \citet{prakash2025lookbacks} describe belief tracking as an address-and-payload lookback, whose vocabulary we borrow. Two studies argue that the post-operation state is not maintained: \citet{tang2026entity} find that neither global nor prior states are decodable and that query-relevant information is aggregated once the query is visible, and \citet{oh2026rebinding} find on the boxes swap task that a swap is a local remapping of the queried box's binding ID at readout, not a global re-encoding. We agree the base model does not retrieve that state from the operation span, and the span nonetheless carries a record a trained reader recovers. 

\paragraph{Reading a frozen cache.} \citet{ding2026scit} introduce source--recipient counterfactuals similar to ours, but apply them to latent chain-of-thought checkpoints, where the reasoning is never emitted, and train nothing. \citet{shih2026steering} argue on the write side that an edit counts as a state change only when a later computation uses the edited value, though their editable register comes from training the model to write its running state. KV-Skill \citep{han2026kvskill} trains a read interface for a frozen backbone, but on a separate residual branch and over a carrier built for the reader, as do gist tokens \citep{mu2023gist} and Patchscopes \citep{ghandeharioun2024patchscopes}. Our transplants are interchange interventions on a frozen writer's cache \citep{geiger2021causal,meng2022rome,geva2023dissecting,zhang2024patching}. None of these combines a masking view over an unmodified cache with a frozen writer and a reader trained in isolation, which is what separates what a cache holds from what the model that wrote it retrieves.

% =====================================================================
\section{limitations}
\label{sec:limitations}
\emph{Seeds and families.} Routing, direct payload and payload localization hold for three reader seeds, but trained routing ($+0.625$ to $+0.942$), the value-matched switch ($0.300$--$0.508$) and swap accuracy (spread $0.58$) vary across seeds, and the held-out footprint rests on two gate-passing readers. Only Mistral-7B among five further families passed our capability gate, and its base model already shows a weak payload. \emph{Reader capacity.} Without a capacity-matched control adapter or a rank sweep, the learned reader may compute over the carrier instead of exposing a pre-existing lookup (\cref{app:extra}). \emph{Scope.} We study single-step records on mostly synthetic tasks; composition across updates remains open, and ToMi and GSM8K test the access gap but not the mechanism. \emph{Precision.} The 8B writer is NF4-quantized, and its carrier results lack a bf16 control.

% =====================================================================
\section{Conclusion}
\label{sec:conclusion}

We froze the model that writes a KV cache and trained only the positions that read it. In controlled state updates, the native model uses the operation span mainly to decide where to read, taking the value from visible text. Isolation training adds direct access to that value, read from the same operand-token record that serves as the address. Whether this extends to longer, composed updates in natural text remains open.

\section*{Reproducibility statement}
All masks are set differences over contiguous instance spans and are checked by assertions in the training loop. Transplants assert identical span positions and prefix lengths. Training schedules are fixed per seed before training, and the central effects were evaluated on unseen instances. Hyperparameters, seeds and full result tables are given in the appendix. Code and data generators will be released.

\section*{AI use statement}
Generative AI assistance was used for writing and debugging experiment and figure code, and, during manuscript preparation, for language editing and LaTeX diagnostics. The authors reviewed all code and are responsible for the experimental design, analyses, claims, citations, and final text.

\bibliography{references}
\bibliographystyle{iclr2027_conference}

\newpage
\appendix
% =====================================================================
% Appendix. Only material cited from the main text. Every number is taken from saved result files.
% =====================================================================
\crefalias{section}{appendix}
\crefalias{subsection}{appendix}

\section*{Appendix contents}
\begin{itemize}[leftmargin=1.2em, itemsep=0pt, topsep=2pt]
\item \Cref{app:training}: training recipes and masks
\item \Cref{app:native}: native use on the boxes task
\item \Cref{app:redirect}: controls for the operation span
\item \Cref{app:carrier}: footprint, routing and payload: full tables
\item \Cref{app:localize}: the native routing record
\item \Cref{app:payloc}: payload localization in trained readers
\item \Cref{app:values}: value boundaries of learned access
\item \Cref{app:gsm8k}: GSM8K
\item \Cref{app:nl}: ToMi and MMLU
\item \Cref{app:extra}: module ablation and model scale
\end{itemize}

% ---------------------------------------------------------------------
\section{Training recipes and masks}
\label{app:training}

\begin{table}[H]
\centering
\small
\resizebox{\textwidth}{!}{\begin{tabular}{@{}lp{0.37\textwidth}p{0.37\textwidth}@{}}
\toprule
 & Boxes & Code \\
\midrule
Backbone & Llama-3.2-1B-Instruct, FP32; also Llama-3.1-8B-Instruct and Mistral-7B-Instruct-v0.3, NF4 (bf16 compute) & Llama-3.1-8B-Instruct, NF4 (bf16 compute); carrier probes also Mistral-7B-Instruct-v0.3, NF4 \\
Reader adapter & \multicolumn{2}{l}{LoRA rank 16, $\alpha=32$, on \code{q/k/v/o} at every layer, applied only to query and answer rows} \\
Optimizer & \multicolumn{2}{l}{AdamW, lr $10^{-4}$, weight decay 0, grad-norm clip 1.0} \\
Steps & 256 & 1024 (carrier readers: 768) \\
One step & one pair, both worlds $\times$ both questions & one item, two queries \\
Scoring & first-token argmax & greedy generation, exact match \\
ISO / ORD & \multicolumn{2}{l}{ISO blocks the whole instance except the operation line; ORD blocks nothing} \\
\bottomrule
\end{tabular}}
\caption{Isolation-training recipes. Both tasks train only a reader adapter; the prefix cache is recomputed at every step by the base model with the adapter disabled, without gradient.}
\end{table}

Masks are set differences over the whole contiguous instance, so separator newlines are blocked with everything else, and the training loop asserts that operation tokens are never blocked and description tokens always are. Every donor differs from its recipient only in assigned values, all values are single tokens (in Mistral-7B, digit tokens of equal count), and span positions and prefix lengths are asserted identical at run time (no family was skipped). Intervals are 95\% bootstraps over items or families with 10{,}000 draws. For Mistral-7B the boxes prompts are re-rendered with the model's own chat template, with the system instruction folded into the first user turn; the first answer token is the word-initial piece of the item name, and all 32 families remain position-aligned for transplants.

\begin{table}[H]
\centering
\small
\resizebox{\textwidth}{!}{\begin{tabular}{@{}llccc@{}}
\toprule
Model & Reader & q0 & q1 & Loss (first 64 $\to$ last 64) \\
\midrule
Llama-3.2-1B & BASE & .000 & .000 & -- \\
 & ORD s1 / s2 / s3 & .031 / .000 / .000 & .031 / .031 / .094 & -- \\
 & ISO s1 / s2 / s3 & .750 / .813 / .781 & .875 / .969 / .938 & $1.90\to0.37$ / $2.00\to0.33$ / $1.98\to0.30$ \\
\midrule
Llama-3.1-8B & BASE & .000 & .000 & -- \\
 & ORD s1 / s2 / s3 & .062 / .031 / .031 & .125 / .031 / .250 & -- \\
 & ISO s1 / s2 / s3 & .906 / .875 / .906 & .938 / .969 / .906 & $1.04\to0.09$ / $0.98\to0.07$ / $0.89\to0.12$ \\
\midrule
Mistral-7B & BASE & .062 & .000 & -- \\
 & ORD s1 / s2 / s3 & .250 / .000 / .000 & .156 / .031 / .000 & -- \\
 & ISO s1 / s2 / s3 & 1.000 / .969 / .969 & .844 / .906 / .781 & $0.34\to0.13$ / $0.61\to0.16$ / $0.51\to0.21$ \\
\bottomrule
\end{tabular}}
\caption{Boxes task, \voponly{}, free-vocabulary accuracy per seed (32 families each; same pairs, schedules and seeds for all models). Loss: mean training loss over the first and last 64 of 256 steps. Mask check (base reader names an operated box under \vopen{} / \voponly{} / operation also blocked): $.81/1.00/.41$, $1.00/1.00/.31$ and $.97/1.00/.50$. $S_{\text{BASE}}$: $1.50$\ci{1.16}{1.86}, $4.24$\ci{3.25}{5.33}, $11.43$\ci{9.49}{13.63}; a matched transplant changes the base reader's answer in 1, 0 and 3 of 32 families.}
\end{table}
\label{tab:boxes}

\paragraph{Training, validation and test splits.} For the code task, training, validation and test items come from three disjoint seeds. The prompt format, the scoring rule and the mask construction were fixed on the validation; the test set is drawn from a seed never used before and is scored by a
script written in advance, with the adapters unchanged. Results are reported on the test set unless marked \emph{validation data}.

% ---------------------------------------------------------------------
\section{Native use on the boxes task}
\label{app:native}

Let $L(\cdot)$ be the logit difference between the donor world's and the recipient's answer, and $S_{\text{BASE}} = [L(\text{donor}) - L(\text{own})]_{\text{PAIR}} - [L(\text{donor}) - L(\text{own})]_{\text{IRRELEVANT}}$ on q0 (base Llama-3.2-1B, \voponly{}, 32 families). $S_{\text{BASE}}=1.496$\ci{1.16}{1.86}, positive in 32/32 families. The logit shift rarely changes behavior: under \voponly{} every untransplanted answer is neither the own nor the donor answer, and a matched transplant turns 1/32 into the donor answer. That family's $S$ ($2.1$) lies inside the range of the others ($+0.08$ to $+4.47$), so $S$ does not predict which families flip, and we report $1.496$ and $1/32$ separately.

% ---------------------------------------------------------------------
\section{Controls for the operation span}
\label{app:redirect}
Validation data, code swap, 8B. \textbf{Null-content donor}: every variable takes a value outside the item's own four (asserted), $n=150$ per query. \textbf{Filler span}: every prompt has a line \code{\# state saved} after the swap, whose K/V attended to strictly more of the prompt than the swap line's; the reader sees only that line.

\begin{table}[H]
\centering
\small
\begin{tabular}{@{}lcccc@{}}
\toprule
Reader & SELF q0/q1 & NULL q0/q1 & FLOOR q0/q1 & Filler span only \\
\midrule
BASE & .240 / .133 & .040 / .053 & .073 / .073 & .060 \\
ISO s3 & .680 / .740 & \textbf{.013 / .027} & .073 / .047 & .055 \\
ISO s4 & .747 / .760 & \textbf{.020 / .040} & .087 / .053 & .055 \\
ORD s3 & .133 / .067 & .040 / .020 & .067 / .047 & .065 \\
\bottomrule
\end{tabular}
\caption{Removing the span's content drops every reader to the floor, and a later span with more context does not substitute for it.}
\end{table}
\label{tab:controls}

Transplanting the recipient's own span back reproduces the untransplanted numbers exactly, so changes under other donors are caused by the replacement.

% ---------------------------------------------------------------------
\section{Footprint, routing and payload: full tables}
\label{app:carrier}

Llama-3.1-8B unless marked Mistral-7B (both NF4), 95\% bootstraps.

\begin{table}[H]
\centering
\small
\begin{tabular}{@{}lcccc@{}}
\toprule
Reader & Donor's realized value & Re-execute on recipient & Other & Donor $-$ recompute \\
\midrule
ISO s3 & .438 & .340 & .198 & $+.098$\ci{+.01}{+.18} \\
ISO s4 & .480 & .282 & .235 & $+.198$\ci{+.12}{+.28} \\
ISO s5 & .475 & .315 & -- & $+.160$\ci{+.07}{+.25} \\
BASE & .013 & .820 & .005 & $-.807$\ci{-.85}{-.77} \\
\bottomrule
\end{tabular}

\vspace{6pt}
\begin{tabular}{@{}lcccc@{}}
\toprule
Worlds ($n=120$, one reader) & \sx{} P(D) & \sx{} P(R) & \sy{} P(D) & \sy{} P(R) \\
\midrule
disjoint: donor value nowhere on screen & .108 & .367 & .200 & .408 \\
exchange: donor value on another line & .421 & .267 & .454 & .338 \\
\bottomrule
\end{tabular}
\caption{Swap under \vopen{}, matched donor, 400 answers per reader. These worlds were built by exchanging slot values, so the donor's value is visible in the recipient and the magnitudes are inflated (bottom).}
\end{table}

\begin{table}[H]
\centering
\small
\resizebox{\textwidth}{!}{\begin{tabular}{@{}lccc@{}}
\toprule
Reader & Loss \sx / \sy / \sz / \sw & Binding \sx & Binding \sy \\
\midrule
\multicolumn{4}{@{}l}{\code{a = b}, Llama-3.1-8B} \\
s1 & 2.34$\to$1.09 / 2.39$\to$1.14 / 3.05$\to$2.50 / 3.06$\to$2.54 & $+.570$\ci{+.49}{+.65} & $+.535$\ci{+.45}{+.61} \\
s2 & 2.65$\to$1.12 / 2.54$\to$1.13 / 3.00$\to$2.39 / 2.92$\to$2.53 & $+.55$ & $+.53$ \\
s3 & 2.73$\to$1.11 / 2.68$\to$1.08 / 3.00$\to$2.53 / 2.95$\to$2.41 & $+.57$ & $+.56$ \\
\midrule
\multicolumn{4}{@{}l}{\code{a = b + 1} (held out), Llama-3.1-8B} \\
s1 & 2.43$\to$1.34 / 2.37$\to$0.99 / 2.96$\to$2.80 / 3.17$\to$2.81 & $+.500$\ci{+.42}{+.57} & $+.650$\ci{+.58}{+.71} \\
s3 & 2.69$\to$0.73 / 2.58$\to$0.53 / 2.99$\to$2.83 / 3.06$\to$2.89 & $+.615$ & $+.675$ \\
s2 (excluded) & 2.66$\to$1.02 / 2.47$\to$0.85 / 3.13$\to$2.76 / 3.09$\to$2.90 & $(+.365)$ & $(+.385)$ \\
\midrule
\multicolumn{4}{@{}l}{\code{a = b + 1} (held out), Mistral-7B} \\
s1 & 0.98$\to$0.67 / 1.03$\to$0.55 / 1.19$\to$1.00 / 1.20$\to$1.11 & $+.315$\ci{+.23}{+.40} & $+.395$\ci{+.32}{+.47} \\
s3 & 1.05$\to$0.70 / 1.03$\to$0.60 / 1.20$\to$0.97 / 1.23$\to$0.96 & $+.255$\ci{+.18}{+.33} & $+.355$\ci{+.28}{+.42} \\
s2 (excluded) & 1.07$\to$0.39 / 1.07$\to$0.40 / 1.12$\to$1.00 / 1.20$\to$0.99 & $(+.305)$ & $(+.290)$ \\
\bottomrule
\end{tabular}}
\caption{Footprint readers. Loss: answer cross-entropy over the first and last 64 steps
per role (chance $2.77$; Mistral losses sit on a digit-split scale). Binding: donor binding
under \voponly{}. A reader that does not reach $.5$ on both addressed roles with nothing
masked is marked \emph{excluded}; its bindings are given in parentheses and not used.}
\end{table}
\label{tab:footprint}

For $a=b+1$, the uninvolved frame variable \sw{} gives binding $.000$ (s1, s3) on Llama-3.1-8B and $.000$\ci{-.04}{+.04} / $+.035$\ci{-.02}{+.09} on Mistral-7B (s1 / s3; base model $+.135$ for \sx{}, $+.100$ for \sy{}, $-.020$ for \sw{}); \sz{} holds the same value in both worlds of a pair, so its binding is zero by construction. The read and write sets and the endpoint were written down before the s1 run. That endpoint, an exact-tuple contrast over the \sx{}, \sy{} and \sw{} queries, was null ($+.010$\ci{-.03}{+.05} under \voponly{}): \sw{} is uninvolved, so it carries no donor--recipient difference and the tuple cannot separate the conditions. The footprint contrast we report, addressed $.588$ vs.\ unaddressed $.060$ ($+.527$\ci{+.47}{+.58}), was chosen afterwards.

\paragraph{Routing and payload probes ($n=120$ per reader).} Each world has four slots, TARGET, OPER, ALT and OTH, and the operation is \code{TARGET = OPER + 1}. Carriers: SELF; D\_SAME (differs only in the operand value); D\_ROLE (same values, reads ALT); UNREL (all values replaced). Line masks are applied at query time.

\paragraph{Other model families.}
The experiment is only meaningful in a family whose base model already performs the held-out operation: if it cannot, there is no pre-existing record for a trained reader to expose. We therefore required a base-model \vopen{} write-target accuracy above $.5$ ($n=40$) before training a second family. Qwen2.5-7B \citep{qwen2025qwen25technicalreport} reaches $.400$, Yi-1.5-9B \citep{ai2024yi} $.500$, DeepSeek-LLM-7B-Chat \citep{deepseekai2024deepseekllmscalingopensource} $.013$ and CodeLlama-13B-Instruct \citep{rozière2024codellamaopenfoundation} $.150$; Mistral-7B-Instruct-v0.3 reaches $.613$ and was trained with three seeds (probes $n=120$, footprint $n=100$ families). The value-matched switch is weak: donor-following barely depends on whether the donor's value is on screen.

\begin{table}[H]
\centering
\small
\resizebox{\textwidth}{!}{\begin{tabular}{@{}lcc@{}}
\toprule
Probe & Trained s1 / s2 / s3 & BASE \\
\midrule
Routing, D\_ROLE $-$ SELF & $+.767$\ci{+.69}{+.84} / $+.458$\ci{+.36}{+.56} / $+.758$\ci{+.68}{+.83} & $+.700$\ci{+.62}{+.78} \\
\midrule
Value-matched switch, donor-following (D\_SAME) & .217 / .325 / .275 & .008 \\
\quad mask ALT, change & $-.067$\ci{-.12}{-.02} / $-.117$\ci{-.17}{-.06} / $-.025$\ci{-.08}{+.02} & $-.008$ \\
\quad mask OPER, change & $+.133$\ci{+.07}{+.20} / $+.058$\ci{+.02}{+.10} / $+.150$\ci{+.06}{+.23} & $+.017$ \\
\quad mask OTH, change & $+.075$ / $-.017$ / $.000$ & .000 \\
Value absent from recipient, donor-following & .183 / .225 / .233 & -- \\
\midrule
Direct payload, SELF & .033 / .033 / .050 & .058 \\
Direct payload, D\_SAME & .300 / .325 / .300 & .192 \\
Direct payload, D\_ROLE & .275 / .283 / .325 & .233 \\
Direct payload, UNREL & .033 / .067 / .058 & .042 \\
Direct payload, D\_SAME, operand query & .325 / .308 / .308 & -- \\
\bottomrule
\end{tabular}}
\caption{Mistral-7B, routing and payload for three trained readers (s1/s2/s3) and the base model ($n=120$ per reader).}
\end{table}

\begin{figure}[H]
\centering
\includegraphics[width=\textwidth]{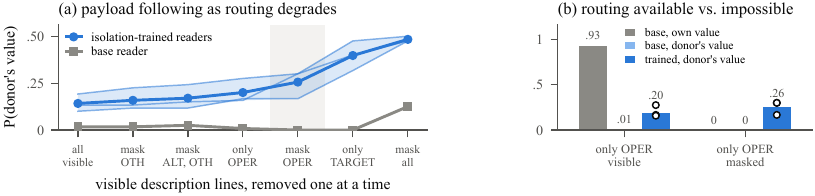}
\vspace{-20pt}
\caption{\textbf{The native payload pathway is absent, not merely unused} (code, Llama-3.1-8B, $n=120$ per reader; D\_SAME donor in the \emph{absent} world, so the donor's value appears nowhere in the recipient). \textbf{(a)} Donor-value rate as visible description lines are removed one at a time; thin lines are the three trained readers, the band their range. \textbf{(b)} The shaded pair of \textbf{(a)}: masking the one line routing needs leaves \emph{more} text visible than masking the other three, yet only the trained readers read more payload there. Circles are seeds.}
\label{fig:ladder}
\end{figure}
% ---------------------------------------------------------------------
\section{The native routing record}
\label{app:localize}

Base models, $n=100$. Bodies consist of six lines of the form \code{name = value \# descriptor}. For sufficiency, the donor's target attributes (name, line position, value, descriptor) are split across recipient variables with distinct values, and each condition moves one attribute. For necessity, we overwrite the recipient's own operand-name token K/V with the state from an identical prompt that reads a dangling name (FRESH), the family mean (MEAN) or zeros (ZERO).

\begin{table}[H]
\centering
\small
\begin{tabular}{@{}lc@{}}
\toprule
Moved attribute & BASE \\
\midrule
name & $+.410$\ci{+.31}{+.51} \\
position only & $+.320$\ci{+.23}{+.41} \\
value & $+.130$\ci{+.07}{+.20} \\
descriptor & $-.010$\ci{-.03}{+.00} \\
\bottomrule
\end{tabular}
\hspace{1em}
\begin{tabular}{@{}lcc@{}}
\toprule
Transplanted part & Joint & Control \\
\midrule
whole span, all layers & $+.460$ & $+.970$ \\
layers 0--3 / 4--7 / 8--11 & .000 / .000 / .000 & .000 \\
\textbf{layers 12--15} & $\mathbf{+.430}$\ci{+.34}{+.53} & $+.950$ \\
layers 16--23 / 24--31 & .000 / .000 & $\le .010$ \\
operand-name token only & $+.340$\ci{+.25}{+.44} & $+.830$ \\
other span tokens only & $+.010$ & $+.090$ \\
\bottomrule
\end{tabular}
\caption{Sufficiency on Llama-3.1-8B. Left: donor-following when one attribute is moved (whole span, all layers). Right: joint condition (all attributes compete) with partial transplants; all-attribute control in the last column. Mistral-7B: \cref{tab:suff-mistral}.}
\end{table}

\begin{table}[H]
\centering
\small
\begin{tabular}{@{}lcc@{}}
\toprule
Transplanted part & Joint & Control \\
\midrule
whole span, all layers & $+.310$\ci{+.22}{+.40} & $+.800$ \\
layers 0--3 / 4--7 / 8--11 & .000 / .000 / $+.010$ & .000 \\
layers 12--15 & $+.060$ & $+.290$ \\
layers 16--23 & $+.130$ & $+.320$ \\
layers 24--31 & .000 & .000 \\
\textbf{layers 14--17} & $\mathbf{+.240}$\ci{+.16}{+.33} & $+.750$ \\
operand-name token only & $+.240$\ci{+.16}{+.33} & $+.740$ \\
other span tokens only & $+.020$ & $+.030$ \\
\bottomrule
\end{tabular}
\caption{Sufficiency on Mistral-7B, joint condition and all-attribute control (the one-attribute conditions were not run). Band 14--17 is added because the record straddles the 4-layer grid.}
\label{tab:suff-mistral}
\end{table}

\begin{table}[H]
\centering
\small
\begin{tabular}{@{}llccc@{}}
\toprule
Model & Band & FRESH & MEAN & ZERO \\
\midrule
Llama-8B & 0--11 & .000 to $-.020$ & .000 & .000 to $-.010$ \\
Llama-8B & \textbf{12--15} & $\mathbf{-.750}$\ci{-.83}{-.66} & $-.540$ & $-.760$ \\
Llama-8B & 16--31 & $\ge -.010$ & $\ge -.010$ & $\ge -.010$ \\
\midrule
Mistral-7B & 0--11 & .000 to $-.010$ & .000 to $-.020$ & .000 to $-.020$ \\
Mistral-7B & 12--15 & $-.150$\ci{-.22}{-.08} & $-.130$ & $-.190$ \\
Mistral-7B & \textbf{16--19} & $\mathbf{-.370}$\ci{-.47}{-.28} & $-.260$ & $-.330$ \\
Mistral-7B & 20--31 & .000 & .000 & .000 \\
Mistral-7B & all 32 & $-.530$\ci{-.63}{-.43} & $-.460$ & $-.610$ \\
\midrule
CodeLlama-13B & \textbf{12--15} & $\mathbf{-.400}$\ci{-.49}{-.31} & $-.300$ & $-.420$ \\
CodeLlama-13B & all 40 & $-.760$ & $-.650$ & $-.730$ \\
\midrule
Qwen-1.5B & 16--19 & $-.320$\ci{-.41}{-.23} & $-.240$ & $-.360$ \\
Qwen-1.5B & all 28 & $-.370$ & $-.330$ & $-.390$ \\
\bottomrule
\end{tabular}
\caption{Necessity at the operand token: change in own-execution accuracy. Llama-3.1-8B (32 layers, own $.990$); Mistral-7B-Instruct (32 layers, own $.660$); CodeLlama-13B-Instruct (40 layers, own $.870$); Qwen2.5-1.5B-Instruct (28 layers, bf16, own $.400$).}
\end{table}

Overwriting the target-name token gives $.000$. On Llama-8B under FRESH at 12--15, answers are the target's old value $+1$ in $.51$ of families: without the record the read defaults to the write target, and the model executes \code{T = T + 1}. On Mistral-7B the record straddles the 4-layer grid (layers 14--17; \cref{app:payloc}), overwriting the target-name token gives $.000$, and the tokens after the operand carry a backup copy (FRESH at all layers, $-.210$), as in CodeLlama-13B.

% ---------------------------------------------------------------------
\section{Payload localization in trained readers}
\label{app:payloc}

The three $a=b+1$ readers of \cref{sec:interfaces} and the base model (Llama-3.1-8B), $n=120$ per reader, all description lines masked. Sufficiency: a D\_SAME donor's K/V is transplanted only at the listed layers and tokens, and we report the donor-value rate minus the no-transplant rate. Necessity: no donor; the reader's own K/V at the listed positions is overwritten, and we report the change in the own-value rate. Token groups: operand name (O), write target (T), the tokens \code{ + 1} (P).

\begin{table}[H]
\centering
\small
\resizebox{\textwidth}{!}{\begin{tabular}{@{}llccc@{}}
\toprule
 & Cell & Trained s1 / s2 / s3 & BASE \\
\midrule
Sufficiency, target query & whole span, all layers & $+.425$ / $+.458$ / $+.483$ & $+.050$ \\
 & operand token, all layers & $+.433$ / $+.467$ / $+.500$ & $+.042$ \\
 & target token / \code{ + 1} / all but operand & $.000$ / $.000$ / $.000$ & $\le +.017$ \\
 & layers 8--11, all tokens & $+.050$ / $+.042$ / $+.117$ & $-.008$ \\
 & layers 12--15, all tokens & $+.325$ / $+.283$ / $+.208$ & $+.042$ \\
 & other six bands & $.000$ / $.000$ / $.000$ & $\le +.017$ \\
 & operand token, layers 12--15 & $+.300$ / $+.275$ / $+.242$ & $+.058$ \\
\midrule
Sufficiency, operand query & whole span, all layers & $+.425$ / $+.442$ / $+.525$ & $+.083$ \\
 & operand token, all layers & $+.425$ / $+.433$ / $+.533$ & $+.067$ \\
 & layers 12--15, all tokens & $+.275$ / $+.375$ / $+.267$ & $+.042$ \\
\midrule
Necessity (intact $.458$ / $.450$ / $.542$) & FRESH, operand, layers 0--7 & $\ge -.008$ & \\
 & FRESH, operand, layers 8--11 & $-.150$ / $-.125$ / $-.225$ & \\
 & FRESH, operand, layers 12--15 & $-.200$ / $-.200$ / $-.175$ & \\
 & FRESH, operand, layers 16--19 & $-.058$ / $-.025$ / $+.025$ & \\
 & FRESH, operand, layers 20--31 & $\ge -.008$ & \\
 & FRESH, operand, all layers & $-.350$ / $-.392$ / $-.425$ & \\
 & FRESH, operand \code{+1}, all layers & $-.392$ / $-.400$ / $-.467$ & \\
 & ZERO, operand, all layers & $-.375$ / $-.425$ / $-.417$ & \\
 & FRESH, target token, all layers & $.000$ / $.000$ / $.000$ & \\
\midrule
Necessity, \vopen{} (intact $.758$ / $.908$ / $.908$) & FRESH, operand, layers 12--15 & $-.500$ / $-.658$ / $-.592$ & \\
 & FRESH, operand, all layers & $-.508$ / $-.800$ / $-.758$ & \\
\bottomrule
\end{tabular}}
\caption{Payload localization, change relative to no transplant (sufficiency) or to the intact carrier (necessity). Per-reader 95\% CIs are about $\pm.09$ for the larger cells.}
\end{table}

\paragraph{Mistral-7B.} The same cells for the three Mistral $a=b+1$ readers and its base model, $n=120$ per reader, no item skipped. Bands 14--17 and 12--19 are added because the record straddles the 4-layer grid.

\begin{figure}[H]
\centering
\includegraphics[width=\textwidth]{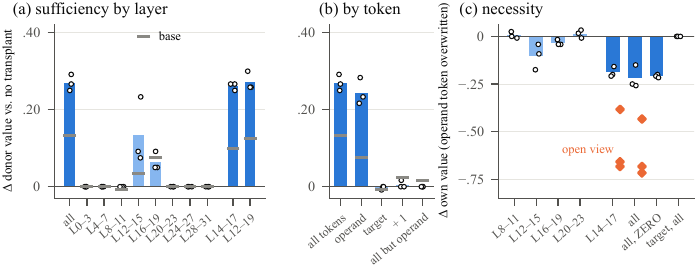}
\vspace{-20pt}
\caption{\textbf{Mistral-7B: the payload is read from the operand-name token in layers 14--17} (all description lines masked, $n=120$ per reader; bars: mean of three trained readers, dots: readers, gray dashes: untrained model). Panels as in \cref{fig:payloc}. The 4-layer grid splits the site between bands 12--15 and 16--19, so panel (a) also shows the bands 14--17 and 12--19.}
\label{fig:payloc-mistral}
\end{figure}

\begin{table}[H]
\centering
\small
\resizebox{\textwidth}{!}{\begin{tabular}{@{}llcc@{}}
\toprule
 & Cell & Trained s1 / s2 / s3 & BASE \\
\midrule
Sufficiency, target query & whole span, all layers & $+.267$ / $+.292$ / $+.250$ & $+.133$ \\
 & operand token, all layers & $+.233$ / $+.283$ / $+.217$ & $+.075$ \\
 & target token / \code{ + 1} / all but operand & $\le +.017$ & $\le +.025$ \\
 & layers 12--15, all tokens & $+.092$ / $+.233$ / $+.075$ & $+.033$ \\
 & layers 16--19, all tokens & $+.050$ / $+.050$ / $+.092$ & $+.075$ \\
 & other six bands & $.000$ / $.000$ / $.000$ & $\le .000$ \\
 & \textbf{layers 14--17, all tokens} & $\mathbf{+.250}$ / $\mathbf{+.267}$ / $\mathbf{+.267}$ & $+.100$ \\
 & layers 12--19, all tokens & $+.258$ / $+.300$ / $+.258$ & $+.125$ \\
 & operand token, layers 14--17 & $+.192$ / $+.242$ / $+.225$ & $+.067$ \\
 & operand token, layers 12--19 & $+.208$ / $+.283$ / $+.233$ & $+.075$ \\
\midrule
Sufficiency, operand query & whole span, all layers & $+.300$ / $+.258$ / $+.283$ & $+.108$ \\
 & operand token, all layers & $+.308$ / $+.242$ / $+.275$ & $+.092$ \\
 & layers 14--17, all tokens & $+.300$ / $+.250$ / $+.283$ & $+.100$ \\
\midrule
Necessity (intact $.275$ / $.308$ / $.317$) & FRESH, operand, layers 0--11 & $\ge -.008$ & \\
 & FRESH, operand, layers 12--15 & $-.042$ / $-.175$ / $-.092$ & \\
 & FRESH, operand, layers 16--19 & $-.017$ / $-.042$ / $-.042$ & \\
 & FRESH, operand, layers 20--31 & $\ge -.008$ & \\
 & \textbf{FRESH, operand, layers 14--17} & $\mathbf{-.158}$ / $\mathbf{-.200}$ / $\mathbf{-.208}$ & \\
 & FRESH, operand, layers 12--19 & $-.158$ / $-.250$ / $-.233$ & \\
 & FRESH, operand, all layers & $-.150$ / $-.250$ / $-.258$ & \\
 & FRESH, operand $+$ \code{ + 1}, all layers & $-.142$ / $-.250$ / $-.275$ & \\
 & ZERO, operand, all layers & $-.200$ / $-.208$ / $-.217$ & \\
 & FRESH, target token, all layers & $.000$ / $.000$ / $.000$ & \\
\midrule
Necessity, \vopen{} (intact $.742$ / $.442$ / $.800$) & FRESH, operand, layers 12--15 & $-.333$ / $-.400$ / $-.217$ & \\
 & FRESH, operand, layers 14--17 & $-.658$ / $-.383$ / $-.683$ & \\
 & FRESH, operand, all layers & $-.683$ / $-.433$ / $-.717$ & \\
\bottomrule
\end{tabular}}
\caption{Mistral-7B payload localization, change relative to no transplant (sufficiency) or to the intact carrier (necessity). Per-reader 95\% CIs are about $\pm.08$ for the larger cells.}
\end{table}
\label{tab:payloc-mistral}

% ---------------------------------------------------------------------
\section{Value boundaries of learned access}
\label{app:values}

\begin{figure}[t]
\centering
\includegraphics[width=\textwidth]{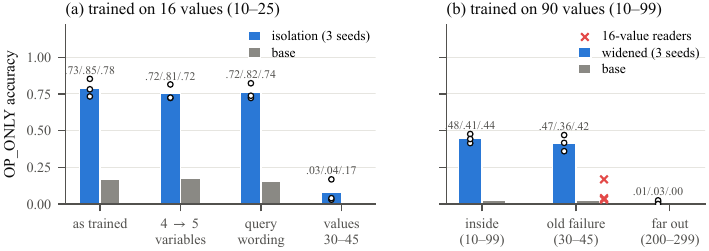}
\vspace{-15pt}
\caption{\textbf{Boundaries of learned access} (code swap, 8B, \voponly{}, $n=400$ pairs per set; dots are seeds). \textbf{(a)} Readers trained on 16 values (s3, s4 and the later s5) survive a change of arity and of wording but collapse on values 30--45, all of which are single tokens. \textbf{(b)} Widening the training values to 10--99 restores accuracy on the old failure set but not 200--299. Crosses: the 16-value readers on 30--45.}
\label{fig:boundary}
\end{figure}

\paragraph{Answer side, not context side.} The value shift changes both the values in context and the value that must be output. A crossed design ($n=400$ items) separates the two: the first letter says whether the other context values are in the trained range (A, 10--25) or far (F, 30--45), the second letter says the same for the answer value.

\begin{table}[H]
\centering
\small
\begin{tabular}{@{}lcccc@{}}
\toprule
Set & BASE & ISO s3 & ISO s4 & ISO s5 (later) \\
\midrule
AA & .188 & .733 & .851 & .787 \\
AF & .000 & \textbf{.041} & \textbf{.049} & \textbf{.180} \\
FA & .142 & .711 & .864 & .757 \\
FF & .000 & .024 & .050 & .166 \\
\bottomrule
\end{tabular}
\caption{Crossed value design, \voponly{}. Accuracy collapses only when the answer is out of range. For BASE, s3 and s4, \vopen{} stays at $.76$--$.99$ in every set.}
\end{table}

\paragraph{Widened training.} Training on \code{range(10, 100)} (90 values, 1024 steps) moves the loss from $4.55$ to $1.44$ (seed 3: $4.38$ to $1.38$).

% ---------------------------------------------------------------------
\section{GSM8K}
\label{app:gsm8k}

Llama-3.2-1B, $n=200$. The writer reads demonstrations, the question and a reference solution. CUT\_PROB\_FIN blocks the question and the last reasoning line for the reader. The gold number appears in a visible intermediate line in 24 of 200 items (copyable); the remaining 176 are clean.

\begin{table}[H]
\centering
\small
\begin{tabular}{@{}lcccc@{}}
\toprule
Reader & Clean 176 & Copyable 24 & All 200 & \vopen{} \\
\midrule
BASE & .023 & .250 & .050 & .970 \\
ORD $\times 3$ & .017/.011/.023 & .250/.250/.292 & .045/.040/.055 & .970/1.000/1.000 \\
ISO $\times 3$ & \textbf{.233/.222/.205} & .292/.208/.292 & .240/.220/.215 & .460/.575/.885 \\
\bottomrule
\end{tabular}
\caption{GSM8K accuracy by stratum under CUT\_PROB\_FIN (first-token argmax), and \vopen{} accuracy.}
\end{table}

Under strict free generation with exact match on clean items, ISO scores $.091/.148/.125$ and ORD $.011/.011/.023$.

% ---------------------------------------------------------------------
\section{ToMi and MMLU}
\label{app:nl}

\paragraph{ToMi.} Llama-3.1-8B, same reader recipe (768 steps). We use stories from the balanced ToMi release with exactly one move line. One training step is a pair of stories whose initial containers differ (same token length) $\times$ \{memory, reality\} questions. Evaluation uses 200 validation memory questions. Under \voponly{}, the move line's K/V are replaced by a matched donor (different initial container) or an irrelevant donor (different room names, same answer). Donor binding is P(follow donor) $-$ P(follow recipient) under the matched donor; selectivity is P(keep recipient $\mid$ irrelevant) $-$ P(keep recipient $\mid$ matched).

\begin{table}[H]
\centering
\small
\begin{tabular}{@{}lccc@{}}
\toprule
Reader & Accuracy & Donor binding & Selectivity \\
\midrule
BASE & .040 & $-.020$ & $+.016$ \\
ORD s1 / s2 / s3 & .015 / .000 / .040 & $-.005$ / $.000$ / $-.005$ & $+.010$ / $+.005$ / $.000$ \\
ISO s1 / s2 / s3 & 1.000 / .970 / .975 & $+.990$ / $+.985$ / $+.990$ & $+1.000$ / $+.979$ / $+.979$ \\
\bottomrule
\end{tabular}
\caption{ToMi memory questions, \voponly{}.}
\end{table}

\paragraph{MMLU.} Base Llama-3.1-8B (NF4), all 1{,}531 validation items, letter-restricted scoring. The question plays the role of the masked description and the answer choices that of the visible span. \vopen{} $.638$; question masked $.599$; question and choices masked $.246$; choices-only prompt, writer never sees the question, $.346$; writer reads a question from another subject, masked $.299$. Question masked minus other question: $+.300$\ci{+.27}{+.33}; on the 1{,}002 items not answerable from the choices alone, $+.310$. The native reader already recovers question information from the choices' cache, so there is no access gap for training to close.

% ---------------------------------------------------------------------
\section{Module ablation and model scale}
\label{app:extra}

\paragraph{Module ablation (code, 8B, \voponly{}, $n=200$, development data).} Untrained LoRA modules keep \code{lora\_B = 0} and are exactly the identity. Training only \code{q} and \code{k} reaches $.705/.800$ (4.2M trainable parameters), only \code{v} and \code{o} reaches $.700/.335$ (6.8M), and all four reaches $.705/.818$ (13.6M); the two halves do not sum because \code{k} and \code{v} are grouped-query projections of lower width. Both halves suffice and neither is necessary. This does not separate redirected  reading from new computation.

\paragraph{Model scale.} Under the same isolation procedure on the code swap task, Llama-3.2-1B stays at the floor, which is why the code experiments use the 8B model.

\end{document}